\documentclass{article}
\usepackage[utf8]{inputenc}
\usepackage{authblk}
\usepackage{setspace}
\usepackage[margin=1.25in]{geometry}
\usepackage{graphicx}
\graphicspath{ {./figures/} }
\usepackage{subcaption}
\usepackage{amsmath}
\usepackage{tabularx}
\usepackage[colorlinks,linkcolor=blue]{hyperref}

\usepackage[style=ieee, 
citestyle=numeric-comp,
sorting=none]{biblatex}
\title{Research on Question Classification Methods in the Medical Field}

\author[1*$\dag$]{Jinzhang LIU}

\affil[1]{Department of Information Engineering, the Chinese University of Hong Kong, Hong Kong, China.}
\affil[*]{Corresponding author. Email: liujinzhang@link.cuhk.edu.hk.com}

\date{}

\begin{document}

\maketitle

\begin{abstract}
Question classification is one of the important links in the research of question and answering system. The existing question classification models are more trained on public data sets. At present, there is a lack of question classification data sets in specific fields, especially in the medical field. To make up for this gap, this paper presents a data set for question classification in the medical field. Moreover, this paper proposes a multi-dimensional extraction of the characteristics of the question by combining multiple neural network models, and proposes a question classification model based on multi-dimensional feature extraction. The experimental results show that the proposed method can effectively improve the performance of question classification.
\end{abstract}


\section{Introduction}
The question answering system is an intelligent system that automatically processes the unstructured text information proposed by the user and returns a unique answer to the user. Question answering system is one of the research directions that have received much attention in the field of natural language processing. The question answering system is mainly composed of three modules: question processing, information retrieval and answer extraction. Problem classification is the basic task of problem processing. After classifying the questions and then performing follow-up operations such as information retrieval, the overall efficiency of the question answering system can be effectively improved \cite{abraham2000ibm}.

Question classification is an operation of classifying the question information of the unstructured text input by the user, which can reduce the number of subsequent retrieved data. Taking the data set used in this article as an example, for the question ``The baby is over 2 years old, has a cold and fever, has a slight cough, should I take medicine?", we classify it according to the departments in the hospital. This question should be in the ``pediatric" category. Similarly, for the question ``reduced menstrual flow, small cysts on the cervix, high serum prolactin", this is a description of gynecology-related diseases, so it should be classified as ``gynecology". After the question is classified, the question answering system only needs to retrieve the information related to the category, which can effectively reduce the computational complexity of information retrieval and greatly improve the overall efficiency of the question answering system.

At present, traditional problem classification methods are based on public datasets. Currently, commonly used public datasets related to problem classification include TREC \cite{voorhees1999trec}, etc. These datasets are wide-ranging datasets. For domain-specific problem classification datasets,
6
 few are published publicly. Especially for datasets in the medical field, it is difficult for us to obtain datasets in the medical field due to the hospital's protection of patient privacy and other reasons. With the development of the Internet, more and more industries have begun to combine with the Internet. For the medical industry, online medical consultation platforms have received more and more attention. Online medical consultation websites provide a platform for doctors and patients to communicate online. At present, the more popular websites in China include Good Doctor, Seeking Doctor and Asking Medicine. This type of website has accumulated a large amount of real medical question and answer information, which provides a wide range of data sources for the research of this paper.

The rest of this paper is organized as follows: Section 1 mainly introduces related work on research on problem classification. Section 2 describes the collection and statistics of the dataset for problem classification in the medical domain. Section 3 describes the research method of problem classification used in the experiments in this paper. Section 4 describes the experimental design and analyzes the experimental results. Section 5 briefly summarizes the conclusions and future work.

\section{Related work}
At present, the experimental methods of problem classification are mainly divided into two types: statistical-based machine learning methods and neural network-based deep learning methods.

\subsection{Statistics-based machine learning methods}
Mishra \cite{mishra2013question} et al. extracted word features, syntactic features, and semantic features from question texts, and trained nearest-neighbor Naive Bayes and SVM classifiers for question classification. Yadav \cite{yadav2013question} et al. used unary, bigram, trigram features and part-of-speech features to classify questions using Naive Bayesian classification. Liu \cite{liu2014chinese} et al. believed that the SVM method of the standard kernel function ignores the structural information of the Chinese question, so they proposed a SMO method of the question text attribute kernel function.

\subsection{neural network-based deep learning methods}
In the field of natural language processing, neural network methods such as Recurrent Neural Networks (RNN), Long Short Term Memory (LSTM) and Gated Recurrent Unit (GRU) are the most widely used currently. In 1990, Elman \cite{elman1990finding} proposed a recurrent neural network framework RNN. Unstructured text is a sequence with sequential order, and RNN can extract sequential information in unstructured text very well. However, RNN has the problem of gradient disappearance and gradient explosion during the training process \cite{hochreiter2001gradient}. In order to solve this problem, in 1997, Hochreiter \cite{hochreiter1997long} and others improved on the RNN structure and proposed a long short-term memory network LSTM. In 2014, Cho \cite{cho2014learning} et al. proposed a variant of the LSTM model, the Gated Recurrent Unit Network GRU. Compared with the LSTM model, the GRU model has a simpler structure and can reduce the time cost of model training.

For many years, the datasets used in traditional problem classification studies are generalized problem types, not domain-specific datasets. In this paper, combined with a large number of real medical question and answer data currently available on the Internet, the crawler technology is used to capture a large number of real and effective medical questions from the medical question and answer website in the past three months, and the captured data is organized into experimental data. The experimental data in this paper have the characteristics of high authenticity, strong real-time performance, and clear domain. The model of the experiment in this paper is based on the current popular neural network technology. Based on the traditional RNN, LSTM and GRU models, the above three models are used to extract features, and then the three features are integrated to form the final feature, and finally the combined features are classified. The model improves the performance of question classification by extracting the features of the question in multiple dimensions.

\section{Dataset}
The medical question answering data used in this article comes from  \href{https://www.120ask.com/list/}{the 120ask question answering community}. This paper captures a total of 17,387 question data in 20 categories. Each piece of data consists of the category of the question and the content of the question. The information about the category and quantity of this data is shown in the Table \ref{tab:1} and the Table \ref{tab:2}.
\begin{table}[h!]
    \caption{Problem category table}    
    \centering
    \begin{tabular}{ccc}
            \hline
Category & Quantity & Proportion\\
\hline
Internal Medicine & 3692 & 21.23\% \\
surgical & 3327 & 19.13\% \\
Obstetrics and Gynecology & 3096& 17.81\%    \\
Pediatrics & 847 & 4.87\%     \\
Traditional Chinese Medicine          & 329      & 1.89\%      \\
Dermatology                           & 2492     & 14.33\%     \\
health care                           & 9        & 0.05\%     \\
Plastic Surgery                       & 263      & 1.51\%     \\
Mental Health Section                 & 511      & 2.94\%     \\
five senses                           & 1222     & 7.03\%     \\
Infectious Diseases                   & 559      & 3.22\%      \\
Oncology                              & 377      & 2.17\%      \\
drug                                  & 12       & 0.07\%    \\
genetic                               & 14       & 0.08\%      \\
home environment                      & 30       & 0.17\%      \\
Other departments                     & 69       & 0.40\%     \\
cosmetic                              & 420      & 2.42\%    \\
Auxiliary examination department      & 22       & 0.13\%     \\
exercise to lose weight               & 61       & 0.35\%     \\
Department of Rehabilitation Medicine & 35       & 0.20\%     \\
            \hline
            \end{tabular}
    \label{tab:1}
\end{table}

\begin{table*}[h!]
    \caption{Partial category data}    
    \centering
    \begin{tabularx}{\textwidth}{lXXX}
            \hline
            category & Question content \\ \hline
            Obstetrics and Gynecology & Decreased menstrual flow, small cysts on the cervix, high serum prolactin \\
            Pediatrics & My baby is more than 2 years old, has a cold and fever, and has a slight cough. Do I need to take medicine?\\
            Internal Medicine & Blood routine check high hemoglobin, is polycythemia vera? \\five senses & Can gums grow back after removal?\\
            \hline
            \end{tabularx}
    \label{tab:2}
\end{table*}

\section{Methods}

In the field of natural language processing, neural network methods such as Recurrent Neural Networks (RNN), Long Short Term Memory (LSTM) and Gated Recurrent Unit (GRU) are currently the most widely used.

\subsection{Common problem classification model}
\subsubsection{Recurrent Neural Networks}
The traditional multi-layer perceptron network assumes that the input data are independent, and the relationship between the data is not input, while in the field of natural language processing, a sentence contains a continuous relationship of sequence. This requires that the input data should contain the temporal relationship within the data. Recurrent Neural Networks are able to capture the internal temporal relationship of the input data.

The RNN unit introduces this relationship with a hidden state or memory, thus preserving the current key information. The hidden state value at any moment is a function of the hidden state value in the previous time step and the input value in the current time step. $t-1$, $t$, $t+1$ represent time series. $x_{t}$ represents the input sample, $s_{t}$ represents the memory of the sample at time $t$, $s_{t}=f\left(U x_{t}+W s_{t-1}\right)$. $W$ represents the weight of the input, $U$ represents the weight of the input sample at this moment, and $V$ represents the weight of the output sample.
At the moment of $t=1$, the input $s_{0}=0$ is generally initialized, $W$, $U$, $V$ are randomly initialized, and then calculated by the following formula:
\begin{equation}
h_{1}=U x_{1}+W s_{0}
\end{equation}
\begin{equation}
s_{1}=f\left(h_{1}\right)
\end{equation}
\begin{equation}
o_{1}=g\left(V s_{1}\right)
\end{equation}
Where $f$ and $g$ are both activation functions, where $f$ can be an activation function such as $tanh$, $relu$, $sigmoid$. $g$ is usually a $softmax$ activation function. When the time advances, the state $s_{1}$ at this time participates in the calculation of time 2 as the memory state of time 1, that is
\begin{equation}
h_{2}=U x_{2}+W s_{1}
\end{equation}
\begin{equation}
s_{2}=f\left(h_{2}\right)
\end{equation}
\begin{equation}
o_{2}=g\left(V s_{2}\right)
\end{equation}
By analogy, the final output can be obtained as:
\begin{equation}
h_{t}=U x_{t}+W s_{t-1}
\end{equation}
\begin{equation}
s_{t}=f\left(h_{t}\right)
\end{equation}
\begin{equation}
o_{t}=g\left(V s_{t}\right)
\end{equation}

\subsubsection{Long Short Term Memory}
The RNN algorithm was introduced earlier. It has a good effect on time series problems, but there are still some problems, among which the more serious problems of gradient disappearance and gradient explosion. Long Short Term Memory (LSTM) is a variant of recurrent neural network that can learn long-term dependencies and can solve the problems of vanishing and exploding gradients. Different from RNN, in RNN is a simple linear summation process, while LSTM achieves the retention of important content and the removal of unimportant content through the ``gate" structure, which solves the gradient disappearance and gradient explosion.

$i$, $f$ and $o$ are input, forget and output gates. As shown below, they use the same formula, but different parameter matrices. The input gate defines how much of the newly calculated state for the current input $x_{t}$ can pass, the output gate defines how much of the current state you want to reveal to the next layer, and the forget gate defines how much of the previous state $h_{t-1}$ you want to pass through. The internal hidden state $g$ is calculated based on the current input $x_t$ and the previous state $h_t-1$. The calculation formula of the $g$ state is the same as that of the RNN, but here it is adjusted by the output generated by the output gate $i$. Given $i$, $f$, $o$, and $g$, we can compute $c_{t}$ from the state at the previous time step $c_{t-1}$. Finally, the hidden state $h_{t}$ is computed by multiplying the memory $c_{t}$ by the output gate $o$.
\begin{equation}
i=\sigma\left(W_{i} h_{t-1}+U_{i} x_{t}\right)
\end{equation}
\begin{equation}
f=\sigma\left(W_{f} h_{t-1}+U_{f} x_{t}\right)
\end{equation}
\begin{equation}
o=\sigma\left(W_{o} h_{t-1}+U_{o} x_{t}\right)
\end{equation}
\begin{equation}
g=\tanh \left(W_{g} h_{t-1}+U_{g} x_{t}\right),
\end{equation}
\begin{equation}
c_{t}=\left(c_{t-1} \otimes f\right) \oplus(g \otimes i)
\end{equation}
\begin{equation}
h_{t}=\tanh \left(c_{t}\right) \otimes o,
\end{equation}

\subsubsection{Gated Recurrent Unit}
Gated recurrent unit network is a variant of LSTM. GRU retains LSTM's resistance to vanishing and exploding gradient problems, but its internal structure is simpler, and the updated hidden state during training requires less computation, so training is faster.

Instead of the input, forget and output gates in the LSTM unit, the GRU unit has two gates: the update gate $z$ and the reset gate $r$. The update gate defines how much of the previous memory is retained, and the reset gate defines how to combine the new input with the previous memory.
\begin{equation}
z=\sigma\left(W_{z} h_{t-1}+U_{z} x_{t}\right)
\end{equation}
\begin{equation}
r=\sigma\left(W_{r} h_{t-1}+U_{r} x_{t}\right),
\end{equation}
\begin{equation}
c=\tanh \left(W_{c}\left(h_{t-1} \otimes r\right)+U_{c} x_{t}\right)
\end{equation}
\begin{equation}
h_{t}=(z \otimes c) \oplus\left((1-z) \otimes h_{t-1}\right)
\end{equation}

\subsection{The experimental model of this paper}
\begin{figure}[h!]
    \centering
    \includegraphics[width=0.5\textwidth]{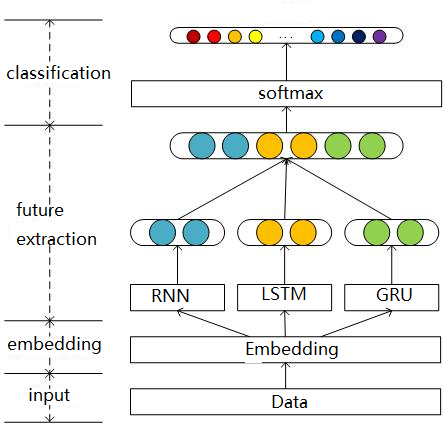}
    \caption{the structure of the problem classification network model}
    \label{fig:1}
\end{figure}

The Figure \ref{fig:1} shows the structure of the problem classification network model proposed in this paper that combines RNN, LSTM and GRU for feature extraction.

In the input layer, we perform a word segmentation operation on a sentence, that is, the input data for training in this paper is a completed sentence of natural language divided into separate words. In the EMBEDDING layer, we used word2vec \cite{mikolov2013distributed} to train word vectors. Then input the RNN structure, LSTM structure and GRU structure respectively for feature extraction. After extracting the features separately, we concatenate the features. Finally, the spliced features are classified by softmax.

\section{Results}
Experimental data: For the Chinese 20 categories of question data used in the laboratory, 60\% are used as the training set, 20\% are used as the validation set, and 20\% are used as the test set. The partitioning of the dataset is random. The specific details of the dataset can be found in Section 3. Parameter setting: The dropout value of all layers in this experiment is set to 0.5, the optimizer uses $adam$, the learning rate is set to 0.01, and the loss function uses $categorical_crossentropy$.Evaluation criteria: The experimental results use the accuracy rate (Accuracy) and the loss value (Loss).

In order to compare with the existing neural network models, we not only implement the method proposed in this paper, but also the traditional RNN neural network model, LSTM neural network model and GRU neural network model.

RNN neural network model, the input layer is the same as the EMBEDDING layer, and only the most basic RNN structure is used for feature extraction in the feature extraction layer.

LSTM neural network model, the input layer is the same as the EMBEDDING layer. In the feature extraction layer, only the most basic LSTM structure is used for feature extraction.

GRU neural network model, the input layer is the same as the EMBEDDING layer. In the feature extraction layer, only the most basic GRU structure is used for feature extraction.
The Table \ref{tab:3} presents the experimental results for different neural network models.
\begin{table}[h!]
    \caption{Experimental result}    
    \centering
    \begin{tabular}{ccc}
            \hline
            Methods    & Accuracy & Loss \\ \hline
            simpleRNN  & 32.63\%  & 2.14\\
            simpleLSTM & 53.39\%  & 1.89\\
            simpleGRU  & 56.50\%  & 1.60\\
            myModel    & 54.00\%  & 1.77\\
            \end{tabular}
    \label{tab:3}
\end{table}
From the experimental results we found that:
Compared with the RNN neural network model, the accuracy of our model is significantly improved by 21.37\%, and the loss value is also reduced by 0.37.
Compared with the LSTM neural network model, our model has a small improvement in accuracy, which is improved by 0.61\%, and the loss value is also reduced by 0.12.
Compared with the GRU neural network model, our model has a lower accuracy rate, a difference of 2.50\%, and the loss value is also higher than the GRU neural network model, which is 0.17 higher.

To sum up, the problem classification network model proposed in this paper that integrates RNN, LSTM and GRU for feature extraction is better than the traditional RNN and LSTM models. The main reason is that our model extracts more features and obtains better results. Much useful information. However, the effect of our experiment is lower than that of the traditional GRU model, the main reason may be that the effect of the RNN model is too poor, which interferes with the extracted final features.

\begin{table*}[h!]
    \caption{Partial category data}    
    \centering
    \begin{tabularx}{\textwidth}{lXXX}
            \hline
category                  & Question content                                                                                              & Problem details                   \\ \hline
Obstetrics and Gynecology & Decreased menstrual flow, small cysts on the   cervix, high serum prolactin                                   & During the Spring Festival, I stay up late, sleep   very lightly, wake up repeatedly, and can't fall asleep. After two or three   months, I always woke up at the same time in the middle of the night. The   menstrual flow suddenly decreased last month, and the menstrual period was   delayed by 4-5 days last week, but this time it is normal.                                                                                                                                                  \\
Pediatrics                & My baby is more than 2 years old, has a cold and   fever, and has a slight cough. Do I need to take medicine? & The baby woke up early this morning and took him   out before six o'clock. It was cloudy and rainy these days and it was cold.   He had a fever of 38.4 in the morning, and the fever was relieved after   taking ibuprofen. He fell asleep at 6 o'clock in the evening and never woke   up. 5. Do you need to use physical cooling? I am afraid of disturbing his   sleep. The current symptoms are fever and occasional coughing. Should I take   medicine or observe it first or recover naturally? \\
Internal Medicine         & Blood routine check high hemoglobin, is   polycythemia vera?                                                  & Disease overview: Sore throat yesterday, I   checked a blood routine with high hemoglobin 195 at a nearby clinic. Disease   details: {[}Male, 28 years old{]} Yesterday, I had a sore throat. I checked a   blood routine with a high hemoglobin 195 at a nearby clinic. There is still   196 in the clinic's blood routine hemoglobin review. Is this polycythemia   vera?                                                                                                                            \\
five senses               & Can gums grow back after removal?                                                                             & Porcelain crown made of six front upper incisors.   The day before yesterday, I went to Dr. Dental Hospital for medicine because   of red, swollen and bleeding gums. The gums in the position of my dentures   were cut. Now I have black triangles on my teeth, which are very ugly. Can   this grow back? ? Can it be recovered?                                     
            \end{tabularx}
    \label{tab:4}
\end{table*}

\section{Discussion}
For the problem classification task, this paper proposes a problem classification method that integrates RNN, LSTM and GRU for feature extraction. The advantage of this method is that it uses features extracted by multiple neural networks. The experimental results show that the method proposed in this paper, in the data used in this paper, compared with the traditional RNN and LSTM models, the classification performance has been greatly improved, which fully shows that the problem classification method proposed in this paper is effective. 

In the next step, we will consider using more classification methods (such as CNN) to further extract effective features. In addition, we also consider using word vectors to replace the word vectors used in this experiment. We will also try to introduce an attention mechanism for experiments to see whether the above methods can effectively improve the performance of problem classification. As shown in the Table \ref{tab:4}, there is still information detailing the problem in our dataset that was not used. We only used the information of the question as input to the experiment. If we combine question data and question detailing data, we can design a question classification model that combines long and short texts. We will try the methods mentioned above in the future to see whether the above methods can effectively improve the performance of problem classification.

\printbibliography

\end{document}